\documentclass{article} 
\usepackage{iclr2024_conference,times}

\usepackage{amsmath,amsfonts,bm}

\def\eqref#1{equation~\ref{#1}}

\def\1{\bm{1}}

\DeclareMathAlphabet{\mathsfit}{\encodingdefault}{\sfdefault}{m}{sl}
\SetMathAlphabet{\mathsfit}{bold}{\encodingdefault}{\sfdefault}{bx}{n}

\DeclareMathOperator*{\argmax}{arg\,max}

\usepackage{hyperref}
\usepackage{url}
\usepackage{graphics} 
\usepackage{epsfig} 
\usepackage{mathptmx} 
\usepackage{times} 
\usepackage{amsmath} 
\usepackage{amssymb}  
\usepackage{graphicx}
\usepackage{mathtools}
\usepackage{bm}
\usepackage{bbold}
\usepackage{pifont}
\usepackage{colortbl}
\usepackage{tabularray}
\usepackage{subcaption}
\usepackage[export]{adjustbox}
\usepackage{float}
\usepackage{multirow}
\usepackage{array}
\usepackage{textcomp}
\usepackage{indentfirst}
\usepackage{booktabs}
\usepackage{bm}
\usepackage{hhline} 
\usepackage{ragged2e}
\usepackage{booktabs}
\usepackage{lipsum}

\newcommand{\cmark}{\ding{51}}%
\newcommand{\xmark}{\ding{55}}%

\newcommand{\inc}[1]{\ensuremath{_{\text{\textcolor{blue}{(+#1)}}}}}

\title{VideoPrompter: An Ensemble of Foundational Models for Zero-Shot Video Understanding}

\author{
  \textbf{Adeel Yousaf}\textsuperscript{1},
  \textbf{Muzammal Naseer}\textsuperscript{2},
  \textbf{Salman Khan}\textsuperscript{2,3},
  \textbf{Fahad Shahbaz Khan}\textsuperscript{2,4},
  \textbf{Mubarak Shah}\textsuperscript{1}
  \\
\hspace{0.9cm}\textsuperscript{1} Center for Research in Computer Vision Lab, University of Central Florida, USA\\
\textsuperscript{2} Mohamed bin Zayed University of AI \hspace{0.15cm}
\textsuperscript{3} Australian National University \hspace{0.15cm}
\textsuperscript{4} Linkoping University
}

\iclrfinalcopy 
\begin{document}

\maketitle

\begin{abstract}


Vision-language models (VLMs) classify the query video by calculating a similarity score between the visual features and text-based class label representations. Recently, large language models (LLMs) have been used to enrich the text-based class labels 
by enhancing the \emph{descriptiveness} of the class names. However, these improvements are restricted to the text-based classifier only, and the query visual features are not considered.
In this paper, we propose 
a framework which combines pre-trained discriminative VLMs with pre-trained generative video-to-text and text-to-text models.  We introduce two key modifications to the standard zero-shot setting. First, we propose language-guided visual feature enhancement and employ a video-to-text model to convert the query video to its descriptive form. The resulting descriptions contain vital visual cues of the query video, such as what objects are present and their spatio-temporal interactions. These descriptive cues provide additional semantic knowledge to VLMs to enhance their zero-shot performance. Second, we propose video-specific prompts to LLMs to generate more meaningful descriptions to enrich class label representations.  Specifically, we introduce prompt techniques to create a Tree Hierarchy of Categories for class names, offering a higher-level action context for additional visual cues, 
We demonstrate the effectiveness of our approach in video understanding across three different zero-shot settings: 1) video action recognition, 2) video-to-text and text-to-video retrieval, and 3) time-sensitive video tasks. Consistent improvements across multiple benchmarks and with various VLMs demonstrate the effectiveness of our proposed framework. Our code will be made publicly available.

\begin{figure}[!htb]
    \makebox[\textwidth][c]{%
        \includegraphics[width=0.95\linewidth]{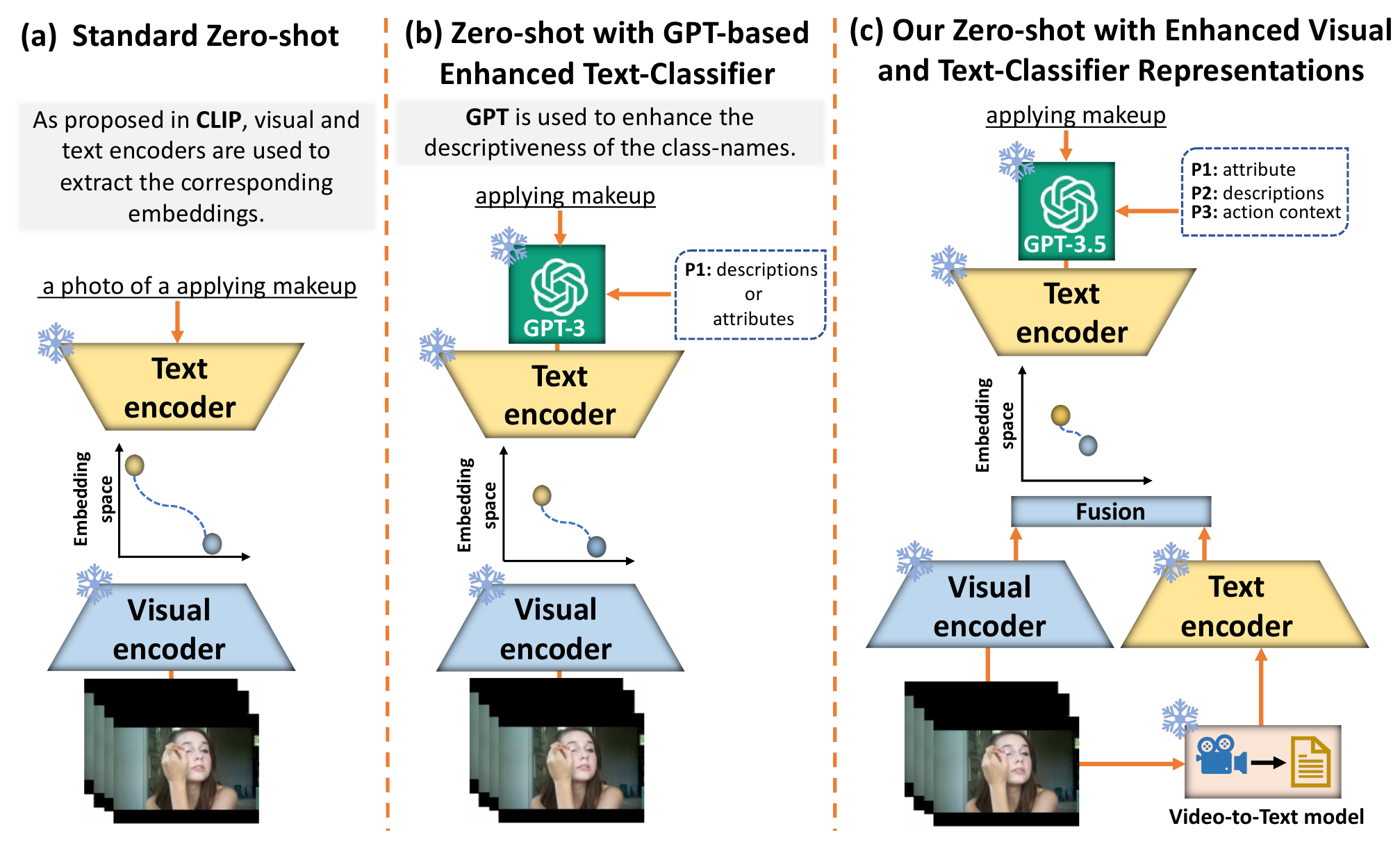}
    }
    \caption{\textbf{(a)} The standard pre-training for zero-shot classification (e.g., CLIP 
    \citep{radford2021learning}). \textbf{(b)} Existing variants for enhancing zero-shot classification \citep{pratt2022does,menon2022visual} using GPT descriptions and attributes that improve text-based classifier features. \textbf{(c)} Our proposed framework to enhance both classifier and visual representations. It employs a video-to-text model to generate description of the query video, and these descriptive cues are combined with the visual information. A text-to-text generative model (GPT-3.5) is prompted for class attributes, descriptions, and action context to enhance the class diversity for the text-based classifier.}
    \label{fig:introfig}
    \vspace{-1.5em}
\end{figure}


\end{abstract}



\section{Introduction}
Open-vocabulary models \citep{roth2023waffling,radford2021learning, jia2021scaling,yuan2021florence} have demonstrated impressive performance in downstream tasks. These models undergo contrastive training on large amounts of image-text pairs, aligning their embeddings in a shared space. However, extending these models to video tasks poses significant challenges mainly for two reasons: (1) due to large computational expenses, and (2) the requirement of gathering large-scale video-text pairs. Therefore, it is critical to effectively leverage pre-trained image-language models for video tasks without affecting their zero-shot capabilities.

To extend pre-trained image-language models to videos, there are two dominant approaches. The first approach takes inspiration from the prompt learning methods \citep{zhou2022learning,zhou2022conditional,jia2022visual} and introduces learnable prompts or adapters to text, vision, or both sides \citep{yang2023aim,wasim2023vita}. In contrast, the second approach fine-tunes the whole pre-trained image-language model for video tasks \citep{Rasheed_2023_CVPR,luo2022clip4clip,wang2021actionclip,ni2022expanding}.  The aforementioned approaches have several drawbacks, e.g., the introduction of additional learnable parameters that add to overall model complexity or require extensive fine-tuning to adapt the model for video tasks. Further, these methods require access to the true distribution of the target task, which can be prohibitive in test-time adaptation and data-scarce environments.

Recently, a new line of work has emerged \citep{menon2022visual,pratt2022does,novack2023chils,roth2023waffling} that incorporates large language models (LLMs), such as GPT-3 to provide additional semantic context to existing vision-language models (VLMs) and requires no further training. These methods query LLMs to replace class names with enriched language descriptors in order to increase class descriptiveness. However, these studies have primarily focused on modifying the text-based classifier only, we propose a twofold improvement approach, simultaneously refining class representations and enriching visual features. Moreover, despite their promising results in image classification, \emph{the applicability of these method in the context of video understanding remains an open question, and our work aims to address this gap}.

We present a framework called VideoPrompter 
that aims to enhance the test-time zero-shot performance of the existing VLMs for video understanding and introduce two modifications to the standard zero-shot framework. \emph{First,} we query a video-to-text generative model to convert the input video to language representation, as this generated representation contains vital visual cues (\emph{such as what objects are present and how they interact spatially and temporally}), which in turn helps the VLMs to better understand the given video. For instance, for a video shown in Figure \ref{fig:introfig}, the video-to-text model can provide detailed textual description e.g., ``\emph{in this video, a woman is seen applying mascara using a makeup brush"}. As humans, we can effortlessly leverage such descriptive information and form a visual image of the video's content in our minds. \emph{Second,} to enrich the class representations of the text classifier, we query LLM with video-specific prompts and generate two types of video-specific descriptors: 1) language attributes and, 2) language descriptions. Moreover, we found that action recognition benchmarks such as UCF-101, HMDB-51 and Kinetics \citep{kuehne2011hmdb, soomro2012ucf101,kay2017kinetics} can be divided into a Tree 
Hierarchy of Categories by using LLM to provide high-level action context. For instance, all sports-related actions (basketball, cricket, baseball) can be added under one high-level action context, i.e., ``playing sports". This high-level action context provides additional cues to further enrich the class representations. For example,  for the action class ``hopscotch", our framework generates following language descriptors:
{\fontfamily{qcr}\selectfont\small
\begin{itemize}\setlength{\itemsep}{0em}
        \item \textbf{parent context:} playing sports.
        \item \textbf{language attributes:} grid markings, hopping, throwing an object, jumping.
        \item \textbf{language descriptions:} a child is playing hopscotch, they hop in a specific pattern, throwing a small object into numbered squares and then taking turns to retrieve it while maintaining balance.
\end{itemize}
}

To summarize, we make the following contributions:
\begin{enumerate}
        \item We introduce a framework that is an ensemble of video-to-text and text-to-text generative models to increase the zero-shot performance of existing VLMs for video understanding.        
        \item We introduce video-specific prompts to enrich the classifier representations and also propose a novel way to generate high-level action contexts from the dataset.
        \item Our framework offers a plug-and-play module adaptable to various existing VLMs such as CLIP \citep{radford2021learning}, ViFi-CLIP \citep{Rasheed_2023_CVPR}, Action-CLIP \citep{wang2021actionclip}, and AIM \citep{yang2023aim}.
        \item We demonstrate results on three different video settings namely: action recognition, video-to-text, and text-to-video retrieval, and time-aware video tasks, and show results and ablation on 7 different datasets.
\end{enumerate}

\section{VideoPrompter}

\subsection{Preliminaries}
Let $\bm{x}$ denote the query video and $C$ denote the target categories. Let $\bm{f}_{V}$ and $\bm{f}_{T}$ respectively be the visual and text encoders of a VLM such as CLIP. The zero-shot video classification can be defined as nearest neighbor retrieval as follows:
\begin{equation}
    \tilde{c} = \argmax_{c\in C} \;\; \cos\left(\bm{f}_{V}(\bm{x}), \bm{f}_{T}(\bm{p}(c))\right),
\end{equation}

with prompt $\bm{p}(c)=$ \texttt A photo of a \{c\},  $\bm cos$ represents cosine similarity between visual and textual features.

Let  $ \bm{F}_{\Theta}$ represent a video-to-text conversational model that converts the input query video $\bm{x}$ to its corresponding video textual description. The generated description embeds vital visual cues about the objects and the action being performed by the objects in the video. This video textual description is then processed by the VLM text encoder, $\bm{f}_{T}$, and fused together with the video features to get the enriched visual features $\bm{\tilde{f}}_V$.

Let  $\bm{F}_\phi$ denote an LLM model that converts the target categories $C$ to corresponding language attributes and descriptions. These language attributes and descriptions provide additional semantic context to enrich class label representations. These class label descriptions are then processed by the VLM text encoder, $\bm{f}_{T}$, to enhance class label features, $\bm{\tilde{f}}_T$. Our proposed 
zero-shot classification can then be defined as follows:
\begin{equation}
    \tilde{c} = \argmax_{c\in C} \;\; \cos (\bm{\tilde{f}}_V , \bm{\tilde{f}}_T\vert_c).
\end{equation}
In the following sections, we examine how the visual features $\bm{\tilde{f}}_V$ and the enriched class representations $\bm{\tilde{f}}_T$ are derived. For clarity, the descriptions generated by the video-to-text model \citep{maaz2023video} are referred to as video textual descriptions. On the other hand, attributes and descriptions generated by LLM \citep{brown2020language} are called language attributes and language descriptions.

\subsection{Video-to-text Guided Visual Feature Enhancement}
\label{subsec:Video-to-text Guided Visual Feature Enhancement}
Given a video-language model, we aim to enhance its zero-shot performance 
by employing a video-to-text conversational model, i.e., Video-ChatGPT (VGPT) \citep{maaz2023video}, to analyze the video content. We prompt VGPT as ``\emph{describe the activity in the video}". VGPT leverages its spatiotemporal alignment between BLIP \citep{li2022blip} and an LLM \citep{chiang2023vicuna} 
to capture temporal dynamics and frame-to-frame consistency relationships, allowing it to generate coherent 
video textual descriptions of the events unfolding in the video. These video textual descriptions embed vital spatial and temporal information about the video events. 
The generated video textual descriptions are passed through the VLM text encoder $\bm{f}_{T}$ to generate a video description-level embedding.

Since the VLMs share a common image-text embedding space due to their contrastive learning objective \citep{radford2021learning}, we found a simple weighted average of the video embedding and video textual description embedding to be efficient. The enhanced visual embedding $\bm{\tilde{f}}_V$ is given as:
\begin{equation}
    \bm{\tilde{f}}_V = \beta_{\substack{1}} \cdot \bm{f}_{V}(\bm{x}) + \beta_{\substack{2}} \cdot \bm{f}_{T}\left(\bm{F}_{\Theta}(x)\right),
\end{equation}


where \(\beta_{\substack{1\sim 2}}\) denotes the weights for the query video and video textual description embeddings, respectively.  


\subsection{Text-to-Text Guided Classifier Refinement}

In our approach, we leverage an LLM, \citep{brown2020language} and generate video-specific descriptors. We found that video-specific descriptors can better adapt to the action-recognition datasets, and unlike \citep{pratt2022does}, only a small number of descriptors are required. For example, our framework only requires three descriptors, unlike 50 for \citep{pratt2022does} in the case of the UCF-101 dataset \citep{soomro2012ucf101} \ref{Comparison with CUPL}.
We arrange video action datasets in high-level action contexts.
For instance, all sports-related actions (basketball, cricket, baseball) can be added under one high-level action context, i.e., ``playing sports". We propose to employ LLM to exploit this property of action-recognition datasets and generate high-level action context to provide additional cues to the classifier. 
In summary, we extract video-specific language descriptions (Sec. \ref{subsubsec:Video-Specific Language Descriptors}) from an LLM with high-level action context (Sec. \ref{subsubsec: Integrating High-Level Video Action Context}). 

\begin{figure}[!t]
    \makebox[\textwidth][c]{%
        \includegraphics[width=1.0\linewidth]{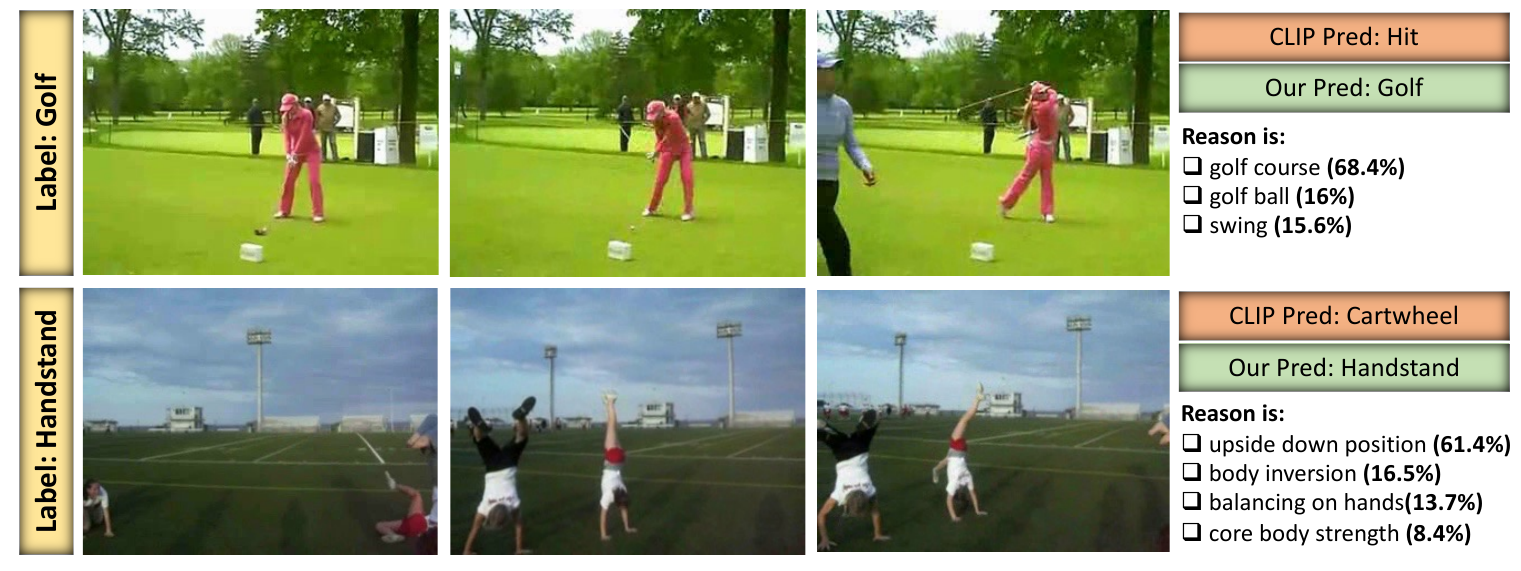}
    }
    \caption{VideoPrompter's Interpretability. We divide our proposed attributes (section \ref{subsubsec:Video-Specific Language Descriptors}) and get the cosine similarity between the individual attribute and query video to find the contribution of each attribute. For example, the top row shows the prediction of our VideoPrompter is ``golf" and the reason the model made this decision is because of \emph{golf-course} in the background, \emph{golf-ball} in the scene, and \emph{swing} of the golf-stick. Similarly, the bottom row shows the \emph{upside down position} played a vital role in the model to make the decision that it's a ``handstand" action.}
    \label{fig:VideoPromper's Interoperability}
    \vspace{-1em}
\end{figure}

\subsubsection{Video-Specific Language Descriptors}
\label{subsubsec:Video-Specific Language Descriptors}
For action recognition, we propose to leverage an LLM, which is GPT-3.5 in our case \citep{brown2020language} to take a class name as input and generate two types of video-specific language descriptors: 
\begin{enumerate}
    \item  \textbf{Language Attributes} offer object-level visual cues to encourage the classifier to employ these features instead of just the class names. For instance, for the video action of ``baby-crawling" the generated language attributes are: \emph{baby, crawling, hand, knees}. We prompt GPT-3.5 as:
     \begin{quotation}
\centering 
\begin{minipage}{0.9\linewidth} 
    {\texttt{Q: What are the distinct visual characteristics to identify a {\{class-name}\} video action?}} 
    \newline
    
\end{minipage}
\end{quotation}
   
    \item  \textbf{Language Descriptions} describe how a specific action is performed. It includes step-by-step directions for that action to facilitate the model's comprehension of the temporal context. For instance, for the video action of ``baby-crawling" the generated language description is: \emph{a baby is seen on all fours, using their arms and knees to move across the floor. They alternate their arms and legs in a crawling motion, exploring their surroundings with curiosity}. We prompt GPT-3.5 as:
     \begin{quotation}
\centering 
\begin{minipage}{0.9\linewidth} 
    {\texttt{Q: How {\{class-name}\} action is performed visually? }} 
\end{minipage}
\end{quotation}
    
\end{enumerate}

The text encoder $\bm{f}_{T}$ extracts the embeddings of such generated language attributes and language descriptions. The modified class label representation  $\bm{\tilde{f}}_T$ 
is the average of the embeddings of language attributes and language descriptions.

In problem setting of video-to-text or text-to-video retrieval, each video is paired with a corresponding caption. We employ GPT-3.5 to take the input caption and generate informative and semantically similar captions. For instance, for an input caption, ``man is giving a review on a vehicle", the generated caption is ``\emph{a person provides feedback on a car}". Similarly, for an input caption ``baseball player hits ball", the generated caption is ``\emph{the ball is hit by a player in a baseball game"}. We prompt GPT-3.5 as:
     \begin{quotation}
\centering 
\begin{minipage}{0.9\linewidth} 
    {\texttt{Q: Given a caption: {\{input caption}\}, generate a visually similar captions.}} 
    
\end{minipage}
\end{quotation}

The text encoder $\bm{f}_{T}$ extracts the embeddings of the generated captions. The modified class label representation  $\bm{\tilde{f}}_T$ is the average of the embeddings of the original and generated captions.


\subsubsection{Integrating High-Level Video Action Context} 
\label{subsubsec: Integrating High-Level Video Action Context}

We proposed a novel way of querying LLMs and divide the dataset into various \emph{high-level action contexts}  such that all video-action classes semantically close to each other are grouped under one high-level action context. We prompt - GPT-3.5 as:
     \begin{quotation}
\centering 
\begin{minipage}{0.9\linewidth} 
    {\texttt{Q: Divide the list of {\{class-names}\} into parent and child classes. Such that actions that are visually similar to each other are in the same group. If the action is not similar to any other action in the list, assign it to others.}} 
    \newline
\end{minipage}
\end{quotation}
As a result, this converts the standard input prompt (\emph{a photo of a {\{class-name}\}}) to high-level action context prompt: \emph{a photo of a {\{high-level context}\} i.e. {\{class-name}\}}. For instance, for the video action of ``\emph{drumming}", the prompt becomes ``\emph{a photo of a playing music i.e., drumming}". The text encoder $\bm{f}_{T}$ extracts the embeddings of the generated high-level action context prompt, which can be averaged with the embeddings of the language attributes and descriptions to create a context-based classifier. 
%
A summary of high-level action context for the different datasets is provided in Table \ref {Table-1}, while the comprehensive list of Tree Hierarchy of Categories is provided in the Appendix \ref{High-level Action Context for HMDB-51} \ref{High-level Action Context for UCF101}, \ref{High-level Action Context for K400}.

\begin{table}[!t]
\caption{A short summary of high-level action context for HMDB-51, UCF-101, and K400 datasets.}
\centering\small
\vspace{-0.5em}
\begin{tabular}{lp{0.7\linewidth}}
\specialrule{1pt}{0pt}{0pt}
Dataset & High-Level Action-Context \\
\midrule

HMDB-51 & self-grooming, physical activities, sports, eating, social interactions, artistic activities, other \\

\hline\\[-3mm]

UCF-101 & self-grooming, playing music, playing sports, exercise and fitness, water activities, household chores, creative activities, other \\

\hline\\[-3mm]

K400 & self-grooming, playing music, playing sports, exercise and fitness, household chores, social interactions, creative activities, transportation activities, water activities, other \\
\specialrule{1pt}{0pt}{0pt}

\end{tabular}
\label{Table-1}
\vspace{-1em}
\end{table}

\section{Experimental Protocols}
\label{headings}
We evaluate the effectiveness of VideoPrompter under three different video zero-shot settings: a) action recognition, b) text-to-video and video-to-text retrieval, and c) time-sensitive video tasks \citep{bagad2023test}. Additionally, we demonstrate that our VideoPrompter provides interpretability of the model decisions (Figure. \ref{fig:VideoPromper's Interoperability}). 
We use the ViT-B/16 backbone and sparsely sample 32 frames as consecutive frames are highly redundant with single-view evaluation \citep{Rasheed_2023_CVPR}. In the case of CLIP, the video embedding is obtained by averaging the frame-level embeddings, \citep{portillo2021straightforward,Rasheed_2023_CVPR}, while for other methods we follow their default settings. 
We use three video textual descriptions from VGPT and only two language descriptors: language attributes and descriptions. To make sure that all three video textual descriptions generated by VGPT are diverse, we set its temperature (likelihood of selecting lower probability tokens) to 0.5., while for GPT-3.5, as we only prompt the model once for each descriptor, we set its temperature to 0.2 to generate more focused and deterministic descriptors. The video textual descriptions generated by VGPT are trimmed to be consistent within the context length of the text encoder \citep{radford2021learning}, and we also apply CLIP-based filtering as a pre-processing step,  discussed in  \ref{subsubsec:Filtering of Visual descriptions} to remove erroneous video textual descriptions. 
We set $\bm{\beta_{\substack{1}}}$  equal to 1.0, and $\bm{\beta_{\substack{2}}}$ is calculated as cosine-similarity between the embeddings of query video and its video textual description. 
This ensures that a video textual description that is consistent with the query video is given higher weight. As AIM \citep{yang2023aim} only comprises a visual encoder, we add a CLIP text encoder for zero-shot analysis. 

%
\begin{table}[!t]
\caption{\small Zero-shot action recognition (top-1 \%) using our VideoPrompter provides consistent improvements across different VLMs and video datasets.}
\vspace{-1em}
\centering\small
\setlength{\tabcolsep}{7pt}
\scalebox{0.9}[0.9]{
\begin{tabular}{lcllll}
\specialrule{0.5pt}{0pt}{0pt}
\hline\\[-3mm]
Method & VideoPrompter  & HMDB  & UCF  & SSv2 & K400 \\ \hline

\multicolumn{6}{c}{\cellcolor[HTML]{EFEFEF} Uni-modal zero-shot action recognition models}                                                \\ \hline
ASR \citep{wang2017alternative}   & --           & 21.8           & 54.4           & --             & --             \\
ZSECOC \citep{qin2017zero}     & --           & 22.6           & 15.1           & --             & --             \\
UR \citep{zhu2018towards}         & --          & 24.4           & 17.5           & --             & -             \\
E2E (\cite{brattoli2020rethinking})   & --                & 32.7           & 48             & --             & --             \\
ER-ZSAR  \cite{chen2021elaborative}   & --           & 35.3           & 51.8           & --             & --             \\ \hline
\multicolumn{6}{c}{\cellcolor[HTML]{EFEFEF}Adapting pre-trained image VL models}                                                         \\ \hline
XCLIP (\cite{ni2022expanding})   & --            & 44.6           & 72.0           & --             & --             \\
A5  (\cite{ju2022prompting})    & --              & 44.3           & 69.3           & --             & --             \\ \hline
 \multirow{2}{*}{CLIP \citep{radford2021learning}} & \xmark        & 37.5           & 61.72          & 2.72          & 44.53         \\
&\cmark & 50.79\inc{13.29}               & 72.77\inc{11.05}               & 4.87\inc{2.15}              & 49.17\inc{4.64}               \\
\hline
\multirow{2}{*}{VIFI-CLIP  \citep{Rasheed_2023_CVPR}} & \xmark     & 51.82          & 77.5           & 4.5           & --             \\
 & \cmark & 57.12\inc{5.30}  & 79.56\inc{2.06}  & 5.40\inc{0.87} & --    \\
\hline
\multirow{2}{*}{AIM \citep{yang2023aim}}  & \xmark           & 51.27          & 72.19          & 4.01          & --             \\
    & \cmark   & 54.37\inc{3.10}          & 78.50\inc{6.31}          & 5.84\inc{1.83}          & --             \\
\hline
\multirow{2}{*}{ActionCLIP \citep{wang2021actionclip}} & \xmark    & 49.20          & 69.52          & 4.42          & --             \\
& \cmark & 51.65\inc{2.45} & 77.07\inc{7.55} & 5.27\inc{0.85} &--   \\
\hline
                    \specialrule{0.5pt}{0pt}{0pt}
\end{tabular}}
\label{Table-2}
\vspace{-1em}
\end{table}

\textbf{Video Action Recognition.} 
Our VideoPrompter achieves consistent improvements across various models and benchmarks, as shown in Table \ref{Table-2}. We observe that when CLIP is incorporated within our framework, it performs on par with the fully-finetuned methods like ViFi-CLIP and ActionCLIP and outperforms adapter-based method AIM.
\emph{Effectiveness of High-Level Action Context in Video Action Recognition:}  We observe that for datasets like UCF-101 \citep{soomro2012ucf101}, HMDB-51 \citep{kuehne2011hmdb} and K400 \citep{kay2017kinetics}, having diverse contexts, our proposed method finds effective and natural high-level action contexts and boost the performance across all methods, shown in Table \ref{table-6}. However, for datasets like SSv2 \citep{goyal2017something}, where class names highly correlate to each other such as \emph{letting something roll along a flat surface} and \emph{letting something roll down a slanted surface}, we found GPT-3.5 divides all of the actions in one class ``manipulating objects" and we observed no improvement with high-level action context.


\textbf{Text-to-Video and Video-to-Text Retrieval.} 
We present recall at a rank (R@K, where k = {1,5})  on 1k-A split-set of the MSR-VTT \citep{xu2016msr} dataset with the CLIP model. In the retrieval setting, for each video, we obtain 10 video textual descriptions using VGPT, and, for every caption we generate two more semantically similar but informative captions using GPT-3.5. As shown in Table \ref{table-3} , our method consistently increases the performance.

\textbf{Time-Sensitive Video Tasks.} \cite{bagad2023test} discussed time awareness in video models and showed that the recent VLMs struggle to understand simple temporal relations such as \emph{before and after}. To show that our VideoPrompter can increase the time-awareness of the existing VLMs, we report the time-consistency score on the before/after synthetic dataset (details of the dataset provided in Appendix \ref{Time-aware synthetic dataset}) \citep{bagad2023test} and time-aware setting of Charades dataset \citep{sigurdsson2016hollywood}.
For each query video, we obtain 10 video textual descriptions using VGPT. In a time-aware setting, considering the nature of the problem, language attributes and descriptions are not used. Our framework enhances the performance on both benchmarks (Table \ref{table-4}). 
We obtain a substantial gain of $10 \%$, even without using the language attributes and descriptions, which shows that our framework can provide additional cues to the VLMs to understand temporal relations.

\begin{table}[!t]
\setlength{\tabcolsep}{5pt}
\caption{\small Our VideoPrompter boosts zero-shot Text-to-Video and Video-to-Text Retrieval performance.}
\vspace{-1em}
\centering
\scalebox{0.9}[0.9]{
\begin{tabular}{lcccccc}
\specialrule{1pt}{0pt}{0pt}
& & & \multicolumn{2}{c}{
  Video-to-Text} &    \multicolumn{2}{c}{Text-to-Video}                          \\ 
  \cmidrule{4-7} 
Method & VGPT & GPT3.5 & R@1    & R@5   & R@1   & R@5    \\ \hline
CLIP  \citep{radford2021learning}        &-- &  --  & 28.19           & 52.90          & 31.7           & 54.0            \\
Video-CLIP \citep{xu2021videoclip} &--  & --      & 30.6            & --               & 10.4           & 22.2            \\
FrozenInTime \citep{bain2021frozen} &--  & --    & --                & --               & 24.7           & 46.9            \\
CLIP4CLIP \citep{luo2022clip4clip} & -- &  --        & --                & --               & 32.0           & \textbf{57.0}   \\ \hline
\multirow{2}{*}{CLIP  \citep{radford2021learning}} & \cmark & \xmark    & 30.59           & 53.90          & 32.8           & 54.5            \\
 & \cmark & \cmark & 31.30\inc{3.11} &
  55.10\inc{2.2} &
  33.50\inc{1.8} &
  56.49\inc{2.49}  \\
  
  \specialrule{1pt}{0pt}{0pt}
\end{tabular}}
\label{table-3}
\vspace{-0.5em}
\end{table}

\begin{table}
\begin{minipage}{0.45\textwidth} 

\setlength{\tabcolsep}{1.0pt}

\caption{\small Our VideoPrompter increases the time awareness in VLMs. SD shows synthetic dataset.}
\vspace{-1em}
\centering
\scalebox{0.68}[0.68]{
\begin{tabular}{lcc}
\specialrule{1pt}{0pt}{0pt}
\multicolumn{1}{l}{Method}  & \multicolumn{2}{c}{Time-consistency score}                                        \\ \hline
                                   & SD                   & Charades                           \\ \cline{2-3} 
CLIP  \citep{radford2021learning}                                  & 50.0 & 56.0 \\
Video-CLIP \citep{xu2021videoclip}      & 51.1 & 47.1 \\
CLIP4Clip \citep{luo2022clip4clip}      & 51.1 & -    \\
CLIP2Video \citep{fang2021clip2video}     & 47.8 & -    \\
CenterCLIP \citep{zhao2022centerclip}     & 46.1 & -    \\
VindLU \citep{cheng2023vindlu}         & 52.0 & -    \\
Frozen in Time \citep{bain2021frozen} & 50.0 & -    \\ \hline
VideoPrompter \hspace{0.01cm}(CLIP)                         & 60.0 \inc{10}  & 57.4 \inc{1.4} \\ 
\specialrule{1pt}{0pt}{0pt}
\end{tabular}}
\label{table-4}
\end{minipage}%
\hfill
\begin{minipage}{0.535\textwidth}
\label{CUPL_comparison}

\setlength{\tabcolsep}{1.5pt}
\vspace{2mm}
\centering
\caption{\small  CUPL generates 50 descriptions for each class. VideoPrompter outperforms CUPL with only 3 language descriptors and a video textual description.} 
\vspace{-1em}
\scalebox{0.7}[0.7]{
\begin{tabular}{lcclllc}
\specialrule{0.5pt}{0pt}{0pt}
\hline
Method    &GPT &VGPT  & HMDB-51 & UCF101  & SSv2 &  Prompts \\
\hline

CLIP          &\-- &\--  & 37.5  & 61.72    & 2.72     & -  \\
\hline
CUPL         &\cmark &\xmark               & 49.14      & 73.67    & 4.06     & 50 \\
CUPL     &\cmark &\cmark               & 50.44      & 73.54    & 4.81     & 50 \\
\hline

VideoPrompter     &\cmark &\cmark                 & 52.51 \inc{3.37} & 73.88\inc{0.21}     & 4.87\inc{0.81}      & 3 \\

\hline
\specialrule{0.5pt}{0pt}{0pt}

\end{tabular}}
\label{table-5}
  \end{minipage}
  \vspace{-0.5em}
\end{table}

\subsection{Understanding different components of VideoPrompter}
In this section, we study different components of our framework. We use the CLIP model with ViT-B/16 visual encoder in all our ablations.

\begin{table}[!t]
\centering
\caption{ \small Action context to further aid action-recognition.}
\vspace{-1em}
\centering\small
\setlength{\tabcolsep}{3pt}

\scalebox{0.95}[0.95]{

\begin{tabular}{llcclll}

\specialrule{0.5pt}{0pt}{0pt}
\hline\\[-3mm]

& Method       & VideoPrompter & Action-Context  & HMDB    & UCF & K400      \\
\hline
& CLIP \citep{radford2021learning}           & \xmark   & \xmark              & 37.5               & 61.72     & 44.53      \\
& CLIP \citep{radford2021learning}           & \cmark   & \xmark            & 50.79              & 72.77     & 49.17      \\
& CLIP \citep{radford2021learning}          & \cmark   & \cmark                & 52.51 \inc{15.01}     &73.88 \inc{12.16}  &52.03 \inc{7.50}    \\

\hline
& ViFi-CLIP \citep{Rasheed_2023_CVPR}         & \xmark   & \xmark            & 51.82              & 77.5    & -         \\
& ViFI-CLIP\citep{Rasheed_2023_CVPR} & \cmark   & \xmark        & 57.12              & 79.56      & -       \\
& ViFI-CLIP\citep{Rasheed_2023_CVPR}    & \cmark   & \cmark          & 57.94 \inc{6.12}     & 80.70 \inc{3.2}  & -    \\
\hline

& AIM \citep{yang2023aim}              & \xmark   & \xmark               & 51.27              & 72.19         & -     \\
& AIM\citep{yang2023aim}                & \cmark   & \xmark             & 54.37                & 78.50  & -   \\
& AIM\citep{yang2023aim}      & \cmark   & \cmark              & 55.54 \inc{4.27}    & 77.90 \inc{5.71}         & -     \\
\hline

& Action-CLIP \citep{wang2021actionclip}                & \xmark   & \xmark   & 49.20              & 69.52         & -     \\
& Action-CLIP \citep{wang2021actionclip} & \cmark   & \xmark   & 51.65              & 77.07         & -     \\
& Action-CLIP \citep{wang2021actionclip}      & \cmark   & \cmark    & 54.50 \inc{5.3}   & 77.47 \inc{7.95}  & -     \\



\hline  \specialrule{0.5pt}{0pt}{0pt}
\end{tabular}}
\label{table-6}
\vspace{-0.5em}
\end{table}


\subsubsection{Design Choices}
To study the design effectiveness of our framework, we discuss various other design choices. First, to show that a combination of query video and its video textual description embedding is the optimal choice, we remove the visual encoder and only take the similarity of the video textual description embedding with the class representations, as shown in Figure \ref{figure-3} (left). We observe that both modalities (visual information and corresponding descriptive information) complement each other and removing visual embedding leads to sub-optimal results.

\begin{figure}[!htb]
    \makebox[\textwidth][c]{%
        \begin{minipage}[c]{0.7\textwidth}
            \includegraphics[width=\linewidth]{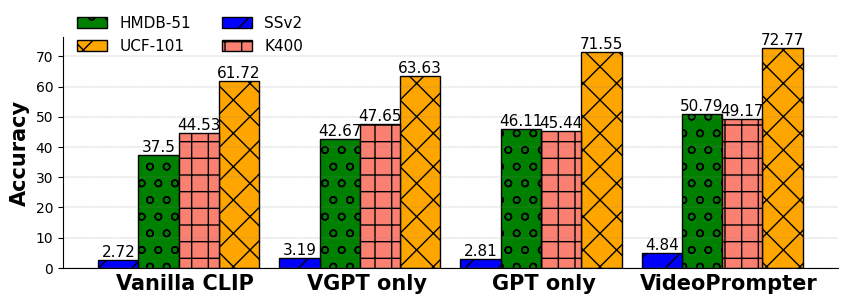}
        \end{minipage}
        \begin{minipage}[c]{0.3\textwidth}
            \vspace{10pt} 
            \caption{Video-textual description and language descriptors complement each other and their combination (VideoPrompter) leads to optimal results. We found this behavior consistent across all benchmarks and models.}
            \label{figure-4}
        \end{minipage}
    }
\end{figure}

Further, we also study the impact of removing either the video-to-text model (VGPT) or the text-to-text model (GPT-3.5), as shown in Figure \ref{figure-4}. While employing these modules individually results in improved performance across all four benchmarks, their combination exhibits a complementary relationship delivering the most optimal performance. 

We also examined the possibility of predicting class names directly by providing GPT-3.5 with video textual descriptions and instructing it to select the closest matching class from a predefined list, we found that this approach fell short of producing optimal results. 

\subsubsection{Filtering of video textual descriptions}
\label{subsubsec:Filtering of Visual descriptions}
We apply CLIP-based filtering as a pre-processing step to remove erroneous visual textual descriptions.  Specifically, we generate 10 visual textual descriptions for each query video and extract corresponding textual embeddings along with the visual embedding of the query video. Cosine similarity between these embeddings is taken to filter the top-3  visual textual descriptions. As shown in Figure \ref{figure-3} (middle), filtering of visual textual descriptions further increases the performance of our framework. We only apply filtering in the action-recognition setting.

\subsubsection{Impact of visual textual description Diversity}

\begin{figure}[!htb]
    \makebox[\textwidth][c]{%
        \includegraphics[width=1.05\linewidth]{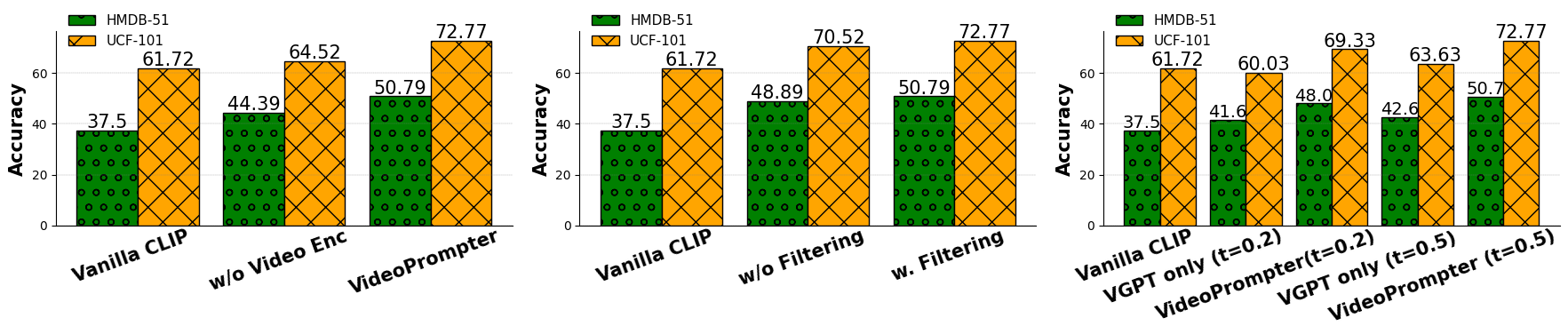}
    }
    \caption{\textbf{(left)} Combination of embeddings of query video and its video textual description leads to optimal choice. \textbf{(middle)} CLIP-based filtering is applied as a pre-processing step, it further boosts the performance by removing the erroneous video textual descriptions. \textbf{(right)} A higher temperature setting leads to better results, as it generates more diverse video textual descriptions.}
    \label{figure-3}
\end{figure}

In order to analyze how the diversity of visual textual descriptions impacts our framework, we experimented with two varying temperature settings (0.2 and 0.5). These temperature settings directly influence the probability of selecting less common tokens, thereby increasing the diversity of the generated descriptions. As shown in Figure \ref{figure-3} (right), the higher temperature setting leads to better results (as it generates more diverse descriptions). Here, \emph{VGPT only} indicate that only video textual descriptions are used, while language attributes and descriptions are not employed.

\subsubsection{Comparison with CUPL}
\label{Comparison with CUPL}
We compare our work with one of the recent works CUPL \citep{pratt2022does} and show that only a handful of carefully designed video prompts combined with the video-to-text guided visual feature enhancement module leads to superior performance. CUPL employs GPT-3 and designs multiple dataset-specific prompts to generate the language descriptions.  For instance, for UCF-101, \citep{pratt2022does} design 5 prompts and generate 10 responses for each prompt leading to 50 descriptions in total. As shown in Table \ref{table-5}, our framework obtains superior performance with only 3 language descriptors i.e. language attributes, language descriptions, and high-level action context.  Further, recent work \citep{roth2023waffling} found descriptor ensemble as the main driver for performance in the case of a large number of prompts and showed comparable performance with randomized descriptors. This further validates our work that only a few carefully designed video prompts are enough to enrich the class representations.

\section{Literature Review}

\textbf{Vision-Language Models.} VLMs \citep{radford2021learning,jia2021scaling} have shown impressive generalization capabilities on various downstream tasks including open-vocabulary image recognition \citep{zhang2021tip,zhou2022learning,zhou2022conditional}, object detection \citep{gu2021open,rao2022denseclip,zhou2022detecting} and image segmentation \citep{ding2022decoupling,zhou2022extract}. 
However, the extension of these models to video-related tasks poses significant challenges, primarily due to the substantial computational costs and the requirement to collect large-scale video-text pairs. Therefore, various methods have been proposed to effectively leverage pre-trained image-language models for video tasks \citep{Rasheed_2023_CVPR,wang2021actionclip, yang2023aim,wasim2023vita}. Nevertheless, these approaches either introduce additional learnable parameters that add to overall model complexity or require extensive fine-tuning to adapt the model for video tasks. Further, these methods require access to the true distribution of the target task, which can be prohibitive in test-time adaptation and data-scarce environments.

\textbf{Improving  Text Classifier Representations Using LLMs.}
As open-vocabulary models classify the input by calculating a similarity score between the image/video and textual prompt (a photo of a {\{class-name}\}.) for each class, this makes the models' performance directly dependent on the \emph{descriptiveness} of the class names. Recently, a new line of work has emerged \citep{menon2022visual,pratt2022does,roth2023waffling,novack2023chils} that incorporates LLMs to enrich these class names and requires no further
training or access to the true distribution of the target task. \citep{menon2022visual} highlights that language can provide additional context to the classifier, and LLM is employed to describe visual features that distinguish that object in an image. \citep{pratt2022does} further explores this idea and uses LLM to generate multiple descriptors for each class name in the dataset. For instance, for UCF-101, \citep{pratt2022does} designed 5 prompts and generated 10 responses for each prompt leading to 50 descriptions in total. Despite the promising results in image classification, \emph{the applicability of these methods \citep{menon2022visual,pratt2022does,roth2023waffling,novack2023chils} in the context of video understanding remains an open question, and our work aims to address this gap}. Moreover, as these methods make use of LLMs to increase the descriptiveness of the class names, the visual side of the VLMs remains unaltered. Our framework - VideoPrompter - simultaneously refines class representations and enriches visual features, utilizing text-to-text  \citep{brown2020language} and video-to-text models \citep{maaz2023video}  respectively. We do so in a two-step approach: first, we query a video-to-text generative model to convert the input video to language representation (description) and fuse it with the query video embedding to enhance the overall visual representation. Second, we use video-specific prompts to query LLM and generate language descriptors to enrich the class representations.

\textbf{Context-based Classifier Enhancement.}
\citep{roth2023waffling, novack2023chils} showed that furnishing contextual information to the VLM can significantly aid the model in directing its attention to relevant features and help resolve class ambiguities. \citep{roth2023waffling} employed LLM  to find the higher-level commonalities in the dataset and GPT-3 is used to extract common context from the datasets. For instance, in CUB200-2011, a bird-related dataset,\citep{roth2023waffling} generates the context ``bird." Likewise, in the eurosat dataset of satellite images, \citep{roth2023waffling} outputs the context ``land use". However,  video-action recognition datasets \citep{kay2017kinetics, kuehne2011hmdb, soomro2012ucf101} generally comprise actions with diverse contexts. For example, the UCF-101 dataset can be subdivided into the following high-level contexts: \emph{self-grooming, playing music, playing sports, exercise and fitness, water activities, household chores, and other activities}, therefore, \citep{roth2023waffling} cannot be directly applied to such diverse datasets. Recently, \citep{novack2023chils} proposed a sub-division of the classes to one level lower, i.e., fine-grained classes. However, their method is only applicable to datasets (unlike action recognition) where the classes are not fine-grained. \emph{We proposed an alternative way of querying LLMs and divide the dataset into various high-level action contexts such that all video-action classes semantically close to each other are grouped under one high-level action context}. For instance, all sports-related actions (basketball, cricket, baseball) can be added under one high-level action context, i.e., ``playing sports". This high-level action context provides additional cues to further enrich the class representations.

\section{Conclusion}
In this work, we introduced a framework - VideoPrompter -  to boost the zero-shot performance of existing VLMs for video understanding. We present a systematic way to prompt pre-trained generative video-to-text and text-to-text models to provide additional semantic context to enrich visual and class representations simultaneously. We demonstrate that, without further training, VideoPrompter performs on par with the various existing fully fine-tuned methods and outperforms adapter-based methods.  We also discuss various design choices and demonstrate that both video-to-text and text-to-text models complement each other and result in optimal performance. We also introduce a Tree Hierarchy of Categories for class names, offering a higher-level action context for additional visual cues. VideoPrompter achieved consistent improvement across three different zero-shot settings: video action recognition, video-to-text, text-to-video retrieval, and time-sensitive video tasks across multiple benchmarks and models.

\section*{Reproducibility}
We used GPT-3.5, CLIP, and VideoChatGPT in our work. All of these models/weights/APIs are publicly available. Additionally, the datasets utilized are also publicly available.  While details to reproduce our work are provided in Section 2, by following the provided instructions, our experiments can be replicated. We will also release the code upon publication.

\bibliography{iclr2024_conference}

\begin{thebibliography}{45}
\providecommand{\natexlab}[1]{#1}
\providecommand{\url}[1]{\texttt{#1}}
\expandafter\ifx\csname urlstyle\endcsname\relax
  \providecommand{\doi}[1]{doi: #1}\else
  \providecommand{\doi}{doi: \begingroup \urlstyle{rm}\Url}\fi

\bibitem[Bagad et~al.(2023)Bagad, Tapaswi, and Snoek]{bagad2023test}
Piyush Bagad, Makarand Tapaswi, and Cees~GM Snoek.
\newblock Test of time: Instilling video-language models with a sense of time.
\newblock In \emph{Proceedings of the IEEE/CVF Conference on Computer Vision and Pattern Recognition}, pp.\  2503--2516, 2023.

\bibitem[Bain et~al.(2021)Bain, Nagrani, Varol, and Zisserman]{bain2021frozen}
Max Bain, Arsha Nagrani, G{\"u}l Varol, and Andrew Zisserman.
\newblock Frozen in time: A joint video and image encoder for end-to-end retrieval.
\newblock In \emph{Proceedings of the IEEE/CVF International Conference on Computer Vision}, pp.\  1728--1738, 2021.

\bibitem[Brattoli et~al.(2020)Brattoli, Tighe, Zhdanov, Perona, and Chalupka]{brattoli2020rethinking}
Biagio Brattoli, Joseph Tighe, Fedor Zhdanov, Pietro Perona, and Krzysztof Chalupka.
\newblock Rethinking zero-shot video classification: End-to-end training for realistic applications.
\newblock In \emph{Proceedings of the IEEE/CVF Conference on Computer Vision and Pattern Recognition}, pp.\  4613--4623, 2020.

\bibitem[Brown et~al.(2020)Brown, Mann, Ryder, Subbiah, Kaplan, Dhariwal, Neelakantan, Shyam, Sastry, Askell, et~al.]{brown2020language}
Tom Brown, Benjamin Mann, Nick Ryder, Melanie Subbiah, Jared~D Kaplan, Prafulla Dhariwal, Arvind Neelakantan, Pranav Shyam, Girish Sastry, Amanda Askell, et~al.
\newblock Language models are few-shot learners.
\newblock \emph{Advances in neural information processing systems}, 33:\penalty0 1877--1901, 2020.

\bibitem[Chen \& Huang(2021)Chen and Huang]{chen2021elaborative}
Shizhe Chen and Dong Huang.
\newblock Elaborative rehearsal for zero-shot action recognition.
\newblock In \emph{Proceedings of the IEEE/CVF International Conference on Computer Vision}, pp.\  13638--13647, 2021.

\bibitem[Cheng et~al.(2023)Cheng, Wang, Lei, Crandall, Bansal, and Bertasius]{cheng2023vindlu}
Feng Cheng, Xizi Wang, Jie Lei, David Crandall, Mohit Bansal, and Gedas Bertasius.
\newblock Vindlu: A recipe for effective video-and-language pretraining.
\newblock In \emph{Proceedings of the IEEE/CVF Conference on Computer Vision and Pattern Recognition}, pp.\  10739--10750, 2023.

\bibitem[Chiang et~al.(2023)Chiang, Li, Lin, Sheng, Wu, Zhang, Zheng, Zhuang, Zhuang, Gonzalez, et~al.]{chiang2023vicuna}
Wei-Lin Chiang, Zhuohan Li, Zi~Lin, Ying Sheng, Zhanghao Wu, Hao Zhang, Lianmin Zheng, Siyuan Zhuang, Yonghao Zhuang, Joseph~E Gonzalez, et~al.
\newblock Vicuna: An open-source chatbot impressing gpt-4 with 90\%* chatgpt quality.
\newblock \emph{See https://vicuna. lmsys. org (accessed 14 April 2023)}, 2023.

\bibitem[Ding et~al.(2022)Ding, Xue, Xia, and Dai]{ding2022decoupling}
Jian Ding, Nan Xue, Gui-Song Xia, and Dengxin Dai.
\newblock Decoupling zero-shot semantic segmentation.
\newblock In \emph{Proceedings of the IEEE/CVF Conference on Computer Vision and Pattern Recognition}, pp.\  11583--11592, 2022.

\bibitem[Fang et~al.(2021)Fang, Xiong, Xu, and Chen]{fang2021clip2video}
Han Fang, Pengfei Xiong, Luhui Xu, and Yu~Chen.
\newblock Clip2video: Mastering video-text retrieval via image clip.
\newblock \emph{arXiv preprint arXiv:2106.11097}, 2021.

\bibitem[Goyal et~al.(2017)Goyal, Ebrahimi~Kahou, Michalski, Materzynska, Westphal, Kim, Haenel, Fruend, Yianilos, Mueller-Freitag, et~al.]{goyal2017something}
Raghav Goyal, Samira Ebrahimi~Kahou, Vincent Michalski, Joanna Materzynska, Susanne Westphal, Heuna Kim, Valentin Haenel, Ingo Fruend, Peter Yianilos, Moritz Mueller-Freitag, et~al.
\newblock The" something something" video database for learning and evaluating visual common sense.
\newblock In \emph{Proceedings of the IEEE international conference on computer vision}, pp.\  5842--5850, 2017.

\bibitem[Gu et~al.(2021)Gu, Lin, Kuo, and Cui]{gu2021open}
Xiuye Gu, Tsung-Yi Lin, Weicheng Kuo, and Yin Cui.
\newblock Open-vocabulary object detection via vision and language knowledge distillation.
\newblock \emph{arXiv preprint arXiv:2104.13921}, 2021.

\bibitem[Jia et~al.(2021)Jia, Yang, Xia, Chen, Parekh, Pham, Le, Sung, Li, and Duerig]{jia2021scaling}
Chao Jia, Yinfei Yang, Ye~Xia, Yi-Ting Chen, Zarana Parekh, Hieu Pham, Quoc Le, Yun-Hsuan Sung, Zhen Li, and Tom Duerig.
\newblock Scaling up visual and vision-language representation learning with noisy text supervision.
\newblock In \emph{International conference on machine learning}, pp.\  4904--4916. PMLR, 2021.

\bibitem[Jia et~al.(2022)Jia, Tang, Chen, Cardie, Belongie, Hariharan, and Lim]{jia2022visual}
Menglin Jia, Luming Tang, Bor-Chun Chen, Claire Cardie, Serge Belongie, Bharath Hariharan, and Ser-Nam Lim.
\newblock Visual prompt tuning.
\newblock In \emph{European Conference on Computer Vision}, pp.\  709--727. Springer, 2022.

\bibitem[Ju et~al.(2022)Ju, Han, Zheng, Zhang, and Xie]{ju2022prompting}
Chen Ju, Tengda Han, Kunhao Zheng, Ya~Zhang, and Weidi Xie.
\newblock Prompting visual-language models for efficient video understanding.
\newblock In \emph{European Conference on Computer Vision}, pp.\  105--124. Springer, 2022.

\bibitem[Kay et~al.(2017)Kay, Carreira, Simonyan, Zhang, Hillier, Vijayanarasimhan, Viola, Green, Back, Natsev, et~al.]{kay2017kinetics}
Will Kay, Joao Carreira, Karen Simonyan, Brian Zhang, Chloe Hillier, Sudheendra Vijayanarasimhan, Fabio Viola, Tim Green, Trevor Back, Paul Natsev, et~al.
\newblock The kinetics human action video dataset.
\newblock \emph{arXiv preprint arXiv:1705.06950}, 2017.

\bibitem[Kuehne et~al.(2011)Kuehne, Jhuang, Garrote, Poggio, and Serre]{kuehne2011hmdb}
Hildegard Kuehne, Hueihan Jhuang, Est{\'\i}baliz Garrote, Tomaso Poggio, and Thomas Serre.
\newblock Hmdb: a large video database for human motion recognition.
\newblock In \emph{2011 International conference on computer vision}, pp.\  2556--2563. IEEE, 2011.

\bibitem[Li et~al.(2022)Li, Li, Xiong, and Hoi]{li2022blip}
Junnan Li, Dongxu Li, Caiming Xiong, and Steven Hoi.
\newblock Blip: Bootstrapping language-image pre-training for unified vision-language understanding and generation.
\newblock In \emph{International Conference on Machine Learning}, pp.\  12888--12900. PMLR, 2022.

\bibitem[Luo et~al.(2022)Luo, Ji, Zhong, Chen, Lei, Duan, and Li]{luo2022clip4clip}
Huaishao Luo, Lei Ji, Ming Zhong, Yang Chen, Wen Lei, Nan Duan, and Tianrui Li.
\newblock Clip4clip: An empirical study of clip for end to end video clip retrieval and captioning.
\newblock \emph{Neurocomputing}, 508:\penalty0 293--304, 2022.

\bibitem[Maaz et~al.(2023)Maaz, Rasheed, Khan, and Khan]{maaz2023video}
Muhammad Maaz, Hanoona Rasheed, Salman Khan, and Fahad~Shahbaz Khan.
\newblock Video-chatgpt: Towards detailed video understanding via large vision and language models.
\newblock \emph{arXiv preprint arXiv:2306.05424}, 2023.

\bibitem[Menon \& Vondrick(2022)Menon and Vondrick]{menon2022visual}
Sachit Menon and Carl Vondrick.
\newblock Visual classification via description from large language models.
\newblock \emph{arXiv preprint arXiv:2210.07183}, 2022.

\bibitem[Ni et~al.(2022)Ni, Peng, Chen, Zhang, Meng, Fu, Xiang, and Ling]{ni2022expanding}
Bolin Ni, Houwen Peng, Minghao Chen, Songyang Zhang, Gaofeng Meng, Jianlong Fu, Shiming Xiang, and Haibin Ling.
\newblock Expanding language-image pretrained models for general video recognition.
\newblock In \emph{European Conference on Computer Vision}, pp.\  1--18. Springer, 2022.

\bibitem[Novack et~al.(2023)Novack, McAuley, Lipton, and Garg]{novack2023chils}
Zachary Novack, Julian McAuley, Zachary~Chase Lipton, and Saurabh Garg.
\newblock Chils: Zero-shot image classification with hierarchical label sets.
\newblock In \emph{International Conference on Machine Learning}, pp.\  26342--26362. PMLR, 2023.

\bibitem[Portillo-Quintero et~al.(2021)Portillo-Quintero, Ortiz-Bayliss, and Terashima-Mar{\'\i}n]{portillo2021straightforward}
Jes{\'u}s~Andr{\'e}s Portillo-Quintero, Jos{\'e}~Carlos Ortiz-Bayliss, and Hugo Terashima-Mar{\'\i}n.
\newblock A straightforward framework for video retrieval using clip.
\newblock In \emph{Mexican Conference on Pattern Recognition}, pp.\  3--12. Springer, 2021.

\bibitem[Pratt et~al.(2022)Pratt, Covert, Liu, and Farhadi]{pratt2022does}
Sarah Pratt, Ian Covert, Rosanne Liu, and Ali Farhadi.
\newblock What does a platypus look like? generating customized prompts for zero-shot image classification.
\newblock \emph{arXiv preprint arXiv:2209.03320}, 2022.

\bibitem[Qin et~al.(2017)Qin, Liu, Shao, Shen, Ni, Chen, and Wang]{qin2017zero}
Jie Qin, Li~Liu, Ling Shao, Fumin Shen, Bingbing Ni, Jiaxin Chen, and Yunhong Wang.
\newblock Zero-shot action recognition with error-correcting output codes.
\newblock In \emph{Proceedings of the IEEE Conference on Computer Vision and Pattern Recognition}, pp.\  2833--2842, 2017.

\bibitem[Radford et~al.(2021)Radford, Kim, Hallacy, Ramesh, Goh, Agarwal, Sastry, Askell, Mishkin, Clark, et~al.]{radford2021learning}
Alec Radford, Jong~Wook Kim, Chris Hallacy, Aditya Ramesh, Gabriel Goh, Sandhini Agarwal, Girish Sastry, Amanda Askell, Pamela Mishkin, Jack Clark, et~al.
\newblock Learning transferable visual models from natural language supervision.
\newblock In \emph{International conference on machine learning}, pp.\  8748--8763. PMLR, 2021.

\bibitem[Rao et~al.(2022)Rao, Zhao, Chen, Tang, Zhu, Huang, Zhou, and Lu]{rao2022denseclip}
Yongming Rao, Wenliang Zhao, Guangyi Chen, Yansong Tang, Zheng Zhu, Guan Huang, Jie Zhou, and Jiwen Lu.
\newblock Denseclip: Language-guided dense prediction with context-aware prompting.
\newblock In \emph{Proceedings of the IEEE/CVF Conference on Computer Vision and Pattern Recognition}, pp.\  18082--18091, 2022.

\bibitem[Rasheed et~al.(2023)Rasheed, Khattak, Maaz, Khan, and Khan]{Rasheed_2023_CVPR}
Hanoona Rasheed, Muhammad~Uzair Khattak, Muhammad Maaz, Salman Khan, and Fahad~Shahbaz Khan.
\newblock Fine-tuned clip models are efficient video learners.
\newblock In \emph{Proceedings of the IEEE/CVF Conference on Computer Vision and Pattern Recognition (CVPR)}, pp.\  6545--6554, June 2023.

\bibitem[Roth et~al.(2023)Roth, Kim, Koepke, Vinyals, Schmid, and Akata]{roth2023waffling}
Karsten Roth, Jae~Myung Kim, A~Koepke, Oriol Vinyals, Cordelia Schmid, and Zeynep Akata.
\newblock Waffling around for performance: Visual classification with random words and broad concepts.
\newblock \emph{arXiv preprint arXiv:2306.07282}, 2023.

\bibitem[Sigurdsson et~al.(2016)Sigurdsson, Varol, Wang, Farhadi, Laptev, and Gupta]{sigurdsson2016hollywood}
Gunnar~A Sigurdsson, G{\"u}l Varol, Xiaolong Wang, Ali Farhadi, Ivan Laptev, and Abhinav Gupta.
\newblock Hollywood in homes: Crowdsourcing data collection for activity understanding.
\newblock In \emph{Computer Vision--ECCV 2016: 14th European Conference, Amsterdam, The Netherlands, October 11--14, 2016, Proceedings, Part I 14}, pp.\  510--526. Springer, 2016.

\bibitem[Soomro et~al.(2012)Soomro, Zamir, and Shah]{soomro2012ucf101}
Khurram Soomro, Amir~Roshan Zamir, and Mubarak Shah.
\newblock Ucf101: A dataset of 101 human actions classes from videos in the wild.
\newblock \emph{arXiv preprint arXiv:1212.0402}, 2012.

\bibitem[Wang et~al.(2021)Wang, Xing, and Liu]{wang2021actionclip}
Mengmeng Wang, Jiazheng Xing, and Yong Liu.
\newblock Actionclip: A new paradigm for video action recognition.
\newblock \emph{arXiv preprint arXiv:2109.08472}, 2021.

\bibitem[Wang \& Chen(2017)Wang and Chen]{wang2017alternative}
Qian Wang and Ke~Chen.
\newblock Alternative semantic representations for zero-shot human action recognition.
\newblock In \emph{Machine Learning and Knowledge Discovery in Databases: European Conference, ECML PKDD 2017, Skopje, Macedonia, September 18--22, 2017, Proceedings, Part I 10}, pp.\  87--102. Springer, 2017.

\bibitem[Wasim et~al.(2023)Wasim, Naseer, Khan, Khan, and Shah]{wasim2023vita}
Syed~Talal Wasim, Muzammal Naseer, Salman Khan, Fahad~Shahbaz Khan, and Mubarak Shah.
\newblock Vita-clip: Video and text adaptive clip via multimodal prompting.
\newblock In \emph{Proceedings of the IEEE/CVF Conference on Computer Vision and Pattern Recognition}, pp.\  23034--23044, 2023.

\bibitem[Xu et~al.(2021)Xu, Ghosh, Huang, Okhonko, Aghajanyan, Metze, Zettlemoyer, and Feichtenhofer]{xu2021videoclip}
Hu~Xu, Gargi Ghosh, Po-Yao Huang, Dmytro Okhonko, Armen Aghajanyan, Florian Metze, Luke Zettlemoyer, and Christoph Feichtenhofer.
\newblock Videoclip: Contrastive pre-training for zero-shot video-text understanding.
\newblock \emph{arXiv preprint arXiv:2109.14084}, 2021.

\bibitem[Xu et~al.(2016)Xu, Mei, Yao, and Rui]{xu2016msr}
Jun Xu, Tao Mei, Ting Yao, and Yong Rui.
\newblock Msr-vtt: A large video description dataset for bridging video and language.
\newblock In \emph{Proceedings of the IEEE conference on computer vision and pattern recognition}, pp.\  5288--5296, 2016.

\bibitem[Yang et~al.(2023)Yang, Zhu, Xie, Zhang, Chen, and Li]{yang2023aim}
Taojiannan Yang, Yi~Zhu, Yusheng Xie, Aston Zhang, Chen Chen, and Mu~Li.
\newblock Aim: Adapting image models for efficient video action recognition.
\newblock \emph{arXiv preprint arXiv:2302.03024}, 2023.

\bibitem[Yuan et~al.(2021)Yuan, Chen, Chen, Codella, Dai, Gao, Hu, Huang, Li, Li, et~al.]{yuan2021florence}
Lu~Yuan, Dongdong Chen, Yi-Ling Chen, Noel Codella, Xiyang Dai, Jianfeng Gao, Houdong Hu, Xuedong Huang, Boxin Li, Chunyuan Li, et~al.
\newblock Florence: A new foundation model for computer vision.
\newblock \emph{arXiv preprint arXiv:2111.11432}, 2021.

\bibitem[Zhang et~al.(2021)Zhang, Fang, Zhang, Gao, Li, Dai, Qiao, and Li]{zhang2021tip}
Renrui Zhang, Rongyao Fang, Wei Zhang, Peng Gao, Kunchang Li, Jifeng Dai, Yu~Qiao, and Hongsheng Li.
\newblock Tip-adapter: Training-free clip-adapter for better vision-language modeling.
\newblock \emph{arXiv preprint arXiv:2111.03930}, 2021.

\bibitem[Zhao et~al.(2022)Zhao, Zhu, Wang, and Yang]{zhao2022centerclip}
Shuai Zhao, Linchao Zhu, Xiaohan Wang, and Yi~Yang.
\newblock Centerclip: Token clustering for efficient text-video retrieval.
\newblock In \emph{Proceedings of the 45th International ACM SIGIR Conference on Research and Development in Information Retrieval}, pp.\  970--981, 2022.

\bibitem[Zhou et~al.(2022{\natexlab{a}})Zhou, Loy, and Dai]{zhou2022extract}
Chong Zhou, Chen~Change Loy, and Bo~Dai.
\newblock Extract free dense labels from clip.
\newblock In \emph{European Conference on Computer Vision}, pp.\  696--712. Springer, 2022{\natexlab{a}}.

\bibitem[Zhou et~al.(2022{\natexlab{b}})Zhou, Yang, Loy, and Liu]{zhou2022conditional}
Kaiyang Zhou, Jingkang Yang, Chen~Change Loy, and Ziwei Liu.
\newblock Conditional prompt learning for vision-language models.
\newblock In \emph{Proceedings of the IEEE/CVF Conference on Computer Vision and Pattern Recognition}, pp.\  16816--16825, 2022{\natexlab{b}}.

\bibitem[Zhou et~al.(2022{\natexlab{c}})Zhou, Yang, Loy, and Liu]{zhou2022learning}
Kaiyang Zhou, Jingkang Yang, Chen~Change Loy, and Ziwei Liu.
\newblock Learning to prompt for vision-language models.
\newblock \emph{International Journal of Computer Vision}, 130\penalty0 (9):\penalty0 2337--2348, 2022{\natexlab{c}}.

\bibitem[Zhou et~al.(2022{\natexlab{d}})Zhou, Girdhar, Joulin, Kr{\"a}henb{\"u}hl, and Misra]{zhou2022detecting}
Xingyi Zhou, Rohit Girdhar, Armand Joulin, Philipp Kr{\"a}henb{\"u}hl, and Ishan Misra.
\newblock Detecting twenty-thousand classes using image-level supervision.
\newblock In \emph{European Conference on Computer Vision}, pp.\  350--368. Springer, 2022{\natexlab{d}}.

\bibitem[Zhu et~al.(2018)Zhu, Long, Guan, Newsam, and Shao]{zhu2018towards}
Yi~Zhu, Yang Long, Yu~Guan, Shawn Newsam, and Ling Shao.
\newblock Towards universal representation for unseen action recognition.
\newblock In \emph{Proceedings of the IEEE conference on computer vision and pattern recognition}, pp.\  9436--9445, 2018.

\end{thebibliography}
\bibliographystyle{iclr2024_conference}

\newpage

\appendix
\section{Time-aware synthetic dataset}
\label{Time-aware synthetic dataset}
The dataset \citep{bagad2023test} contains 180 video-text pairs with different shapes and colors and two types of temporal relationships: \emph{before and after}. Each video has a corresponding correct caption referred to as an \emph{attractor} and a negative caption with flipped events referred to as \emph{distractor}. In the ideal scenario, the VLM should be able to associate query video with the attractor. 

\section{High-level Action Context for HMDB-51}
\label{High-level Action Context for HMDB-51}
\fontfamily{qcr}\selectfont
\textbf{Self-grooming:} \lowercase{brush hair.}\\
\textbf{Physical activities:} \lowercase{Cartwheel, Climb, Climb stairs, Dive, Flicflac, Hand stand, Jump, Pullup, Pushup, Ride bike, Ride horse, Run, Situp, Somersault, Stand, Swing baseball, Talk, Turn, Walk.}\\
\textbf{Sports:} \lowercase{Catch, Dribble, Golf, Hit, Kickball, Kick, Shootball.}\\
\textbf{Eating:} \lowercase{Chew, Drink, Eat.}\\
\textbf{Social interactions:} \lowercase{Clap, Hug, Kiss, Shake hands, Smile, Wave.}\\
\textbf{Artistic activities:} \lowercase{Draw sword, Sit, Smoke, Fencing, Laugh, Pour, Pick, Shoot gun, Shoot bow, Sword exercise, Throw, Sword.}\\
\textbf{Others:} \lowercase{fall floor, push, punch.}

\section{High-level Action Context for UCF101}
\label{High-level Action Context for UCF101}
\fontfamily{qcr}\selectfont

\textbf{Self-grooming:}  \lowercase{Apply Eye Makeup, Apply Lipstick, Blow Dry Hair, Brushing Teeth, Haircut, Head Massage, Shaving Beard.}\\
\textbf{Playing music:} \lowercase{Drumming, Playing Cello, Playing Daf, Playing Dhol, Playing Flute, Playing Guitar, Playing Piano, Playing Sitar, Playing Tabla, Playing Violin.}\\
\textbf{Playing sports:} \lowercase{Archery, Balance Beam, Band Marching, Baseball Pitch, Basketball, Basketball Dunk, Bench Press, Biking, Billiards, Bowling, Boxing Punching Bag, Boxing Speed Bag, Cricket Bowling, Cricket Shot, Field Hockey Penalty, Frisbee Catch, Gold Swing, Hammer Throw, Horse Race.}\\
\textbf{Exercise and fitness:} \lowercase{Body Weight Squats, Handstand Pushups, Pull Ups, Lunges, Handstand Walking, High Jump, Push Ups, Wall Pushups.}\\
\textbf{Water activities:} \lowercase{Breast Stroke, Cliff Diving, Diving, Kayaking, Rafting.}\\
\textbf{Household chores:} \lowercase{Cutting In Kitchen, Mixing, Mopping Floor.}\\ 
\textbf{Creative activities:} \lowercase{Knitting, Typing, Writing On Board, YoYo.}\\
\textbf{Other:} \lowercase{Baby Crawling, Blowing Candles, Hula Hoop, Nunchucks, Parallel Bars, Pizza Tossing, Rope Climbing, Salsa Spin, Swing, Tai Chi, Walking With Dog, Clean And Jerk, Fencing, Front Crawl, Floor Gymnastics, Hammering, Juggling Balls, Jump Rope, Jumping Jack, Pommel Horse, Punch, Sky Diving.}

\section{High-level Action Context for K400}
\label{High-level Action Context for K400}
\fontfamily{qcr}\selectfont

\textbf{Self-grooming:} applying cream, brushing hair, brushing teeth, cutting nails, dying hair, fixing hair, filling eyebrows, getting a haircut, getting a tattoo, grooming dog, grooming horse, massaging back, massaging feet, massaging legs, massaging person's head, shaving head, shaving legs, shining shoes, trimming or shaving beard, waxing back, waxing chest, waxing eyebrows, waxing legs. \\
\textbf{Playing music:} air drumming, beatboxing, playing accordion, playing bagpipes, playing bass guitar, playing cello, playing clarinet, playing controller, playing didgeridoo, playing drums, playing flute, playing guitar, playing harmonica, playing harp, playing ice hockey, playing keyboard, playing organ, playing piano, playing recorder, playing saxophone, playing trombone, playing trumpet, playing ukulele, playing violin,  playing xylophone, strumming guitar, tapping guitar.\\
\textbf{Playing sports:} archery, arm wrestling, bobsledding, bowling, capoeira, cartwheeling, cheerleading, climbing a rope, climbing tree, contact juggling, disc golfing, dodgeball, drop kicking, golf chipping, golf driving, golf putting, high jump, high kick, hitting baseball, hockey stop, hopscotch, hurdling, hurling (sport), ice climbing, javelin throw, kitesurfing, long jump, paragliding, parkour, passing American football (in game), pas sing American football (not in game), picking fruit, playing basketball, playing cricket, playing kickball, playing squash or racquetball, playing tennis, playing olleyball, pole vault, riding mechanical bull, riding mountain bike, roller skating, shooting basketball, shooting goal (soccer), shot put, skiing (not slalom or crosscountry), skiing crosscountry, skiing slalom, skydiving, slacklining, sled dog racing, snowboarding, snowkiting, snowmobiling, somersaulting, spinning poi, springboard diving, swing dancing, sword fighting,tobogganing, trapezing, triple jump, vault, wrestling.\\
\textbf{Exercise and fitness:} abseiling, bench pressing, blasting sand, blowing leaves, blowing nose, blowing out candles, bouncing on trampoline, braiding hair, bungee jumping, carrying baby, chopping wood, clapping, climbing ladder, crawling baby, crossing river, crying, digging, exercising arm, exercising with an exercise ball, extinguishing fire, feeding birds, feeding fish, feeding goats, front raises, garbage collecting, gargling, hammer throw, holding snake, jogging, jumping into pool, laughing, laying bricks, lunge, mopping floor, moving furniture, mowing lawn,  opening present, planting trees, reading book, rock climbing, running on treadmill, scrambling eggs, shaking hands, shaking head, sharpening pencil, shoveling snow, side kick, situp, skipping rope, smoking, smoking hookah, snatch weight lifting, sneezing, sniffing, squat, stomping grapes, stretching arm, stretching leg, surfing crowd, swinging legs, swinging on something, tai chi, taking a shower, tying bow tie, tying knot (not on a tie), tying tie, using remote controller (not gaming), walking the dog, washing feet, washing hair, washing hands, welding, yawning. \\
\textbf{Household chores:} assembling computer, breading or breadcrumbing, building cabinet, building shed, cleaning floor, cleaning gutters, cleaning pool, cleaning shoes, cleaning toilet, cleaning windows, counting money, cutting pineapple, cutting watermelon, doing laundry, doing nails, folding clothes, folding napkins, folding paper, frying vegetables, ironing, making bed, plastering, reading newspaper, ripping paper, sanding floor, setting table, sharpening knives, shredding paper, sweeping floor, trimming trees, unboxing, using computer, washing dishes, watering plants.\\
\textbf{Social interactions:} answering questions, auctioning, celebrating, checking tires, giving or receiving award, news anchoring, presenting weather forecast, testifying, texting, waiting in line.\\
\textbf{Creative activities:} arranging flowers, balloon blowing, bandaging, bartending, bee keeping, blowing glass, bookbinding, carving pumpkin, country line dancing, cracking neck, drawing, making jewelry, playing cards, playing monopoly, recording music, sticking tongue out, weaving basket, wrapping present, writing. \\
\textbf{Transportation activities:} biking through snow, driving car, driving tractor, flying kite, hoverboarding, motorcycling, riding a bike, riding camel, riding elephant, riding mule, riding or walking with horse, riding scooter, riding unicycle, unloading truck, using segway.\\
\textbf{Water activities:} canoeing or kayaking, diving cliff, ice fishing, ice skating, jetskiing, parasailing, sailing, scuba diving, snorkeling, surfing water, swimming breast stroke, swimming butterfly stroke, water skiing, water sliding, windsurfing.\\
\textbf{Other:} baby waking up, bending back, bending metal, brush painting, catching fish, catching or throwing baseball, catching or throwing frisbee, catching or throwing softball, changing oil, changing wheel, clay pottery making, clean and jerk, curling hair, deadlifting, doing aerobics, dribbling basketball, drinking, drinking beer, drinking shots, drumming fingers, dunking basketball, egg hunting, flipping pancake, grinding meat, gymnastics tumbling, hugging, jumpstyle dancing, kicking field goal, kicking soccer ball, kissing, marching, petting animal (not cat), petting cat, playing badminton, playing chess, playing cymbals, playing paintball, playing poker, pull ups, pumping fist, pumping gas, punching bag, punching person (boxing), push up, pushing car, pushing cart, pushing wheelchair, shearing sheep, shuffling cards,  sign language interpreting, ski jumping, slapping, spraying, swimming backstroke, tango dancing, tickling, tossing  coin, training dog.

\end{document}